\documentclass{article} 
\usepackage{nips13submit_e, times, amsthm, amsmath, bbm, amsfonts, amssymb, cite, graphicx, caption, enumerate, stmaryrd, ctable, subfigure, xspace, algorithm, algorithmic, soul, placeins}
\usepackage{hyperref,cancel}

\newcommand{\cO}{{\mathcal O}}
\newcommand{\x}{{\mathbf x}}

\newcommand{\z}{{\mathbf z}}

\newcommand{\vt}{\mathbf{v}}
\newcommand{\wt}{\mathbf{w}}

\newcommand{\bE}{{\mathbb E}}
\newcommand{\bR}{{\mathbb R}}

\newcommand{\cksny}{\texttt{XNV}\xspace}

\newcommand{\Kt}{{\mathbf{K}}} 
\newcommand{\Dt}{{\mathbf{D}}} 
\newcommand{\wccaj}{{{\beta}}}
\newcommand{\wcca}{{\boldsymbol{\wccaj}}}
\newcommand{\Bt}{{\mathbf{B}}} 
\newcommand{\Vt}{{\mathbf{V}}} 

\newcommand{\sq}[1]{\left[ {#1} \right]}
\newcommand{\tr}{^{\top}}

\renewcommand{\algorithmicrequire}{\textbf{Input:}}
\renewcommand{\algorithmicensure}{\textbf{Output:}}

\DeclareMathOperator{\lloss}{loss}
\DeclareMathOperator*{\argmin}{argmin}
\DeclareMathOperator*{\argmax}{argmax}

\newtheorem{thm}{Theorem}

\newtheorem{defn}{Definition}

\title{Randomized co-training: from cortical neurons\\ to machine learning and back again}

\author{David Balduzzi\\
Department of Computer Science\\
ETH Zurich, Switzerland\\
\texttt{david.balduzzi@inf.ethz.ch}}

\nipsfinalcopy 

\begin{document}
\maketitle

\begin{abstract} 
	Despite its size and complexity, the human cortex exhibits striking anatomical regularities, suggesting there may simple meta-algorithms underlying cortical learning and computation. We expect such meta-algorithms to be of interest since they need to operate quickly, scalably and  effectively with little-to-no specialized assumptions. 
	
	This note focuses on a specific question: How can neurons use vast quantities of \emph{unlabeled} data to speed up learning from the comparatively rare labels provided by reward systems? As a partial answer, we propose randomized co-training as a biologically plausible meta-algorithm satisfying the above requirements. As evidence, we describe a biologically-inspired algorithm, Correlated Nystr\"om Views (\cksny) that achieves state-of-the-art performance in semi-supervised learning, and sketch work in progress on a neuronal implementation. 
\end{abstract}

Although staggeringly complex, the human cortex has a remarkably regular structure \cite{douglas:04, douglas:07}. For example, even expert anatomists find it difficult-to-impossible to distinguish between slices of tissue taken from, say, visual and prefrontal areas. This suggests there may be simple, powerful meta-algorithms underlying neuronal learning.

Consider one problem such a meta-algorithm should solve: taking advantage of massive quantities of unlabeled data. Evolution has provided mammals with neuromodulatory systems, such as the dopamenergic system, that assign labels (e.g. pleasure or pain) to certain outcomes. However, these labels are rare; an organism's interactions with its environment are typically indifferent. Nevertheless, mammals often generalize accurately from just a few good or bad outcomes. Our problem is therefore to understand how organisms, and more specifically individual neurons, use unlabeled data to learn quickly and accurately.

Next, consider some properties a semi-supervised \emph{neuronal} learning algorithm should have. It should be: 
\begin{itemize}
	\item fast;
	\item scalable; 
	\item effective; 
	\item broadly applicable (that is, requiring few-to-no specialized assumptions); and
	\item biologically plausible.
\end{itemize}
The first four requirements are of course desirable properties in any learning algorithm. The fourth requirement is particularly important due to the wide range of environments, both stochastic and adversarial, that organisms are exposed to.

Regarding the fifth requirement, it is unlikely that evolution has optimized all of the cortex's $\approx\hspace{-1mm}10^{15}$ connections, especially given the explosive growth in brain size over the last few million years. A simpler explanation, fitting neurophysiological evidence, is that the macroscopic connectivity (largely the white matter) was optimized, and the details are filled in randomly.

\paragraph{Contribution.}
This note proposes \emph{randomized co-training} as a semi-supervised meta-algorithm that satisfies the five criteria above. In particular, we argue that randomizing cortical connectivity is not only necessary but also beneficial.

\paragraph{Co-training.}
A co-training algorithm takes labeled $(x^{(1)}_i, x^{(2)}_i, y_i)_{\iota=1}^\ell$ and unlabeled data $(x^{(1)}_i, x^{(2)}_i)_{i=\ell+1}^{\ell+u}$ consisting of two views \cite{blum:98}. Examples of views are audio and visual recordings of objects or photographs taken from different angles. The key insight is that good predictors will agree on both views whereas bad predictors may not \cite{balcan:10}. Co-training algorithms therefore use unlabeled data to eliminate predictors disagreeing across views, shrinking the search space used on the labeled data -- resulting in better generalization bounds and improved empirical performance \cite{sindhwani:05, farquhar:05, brefeld:06, rosenberg:07, kakade:07, sridharan:08, mbb:13}.

The most spectacular application of co-training is perhaps never-ending language learning (NELL), a semi-autonomous agent that updates a massive database of beliefs about English language categories and relations based on information extracted from the web \cite{carlson:10, carlson:10a, balcan:13}. 

However, despite its conceptual elegance, co-training remains a niche method. A possible reason for the lack of applications is that it is difficult to find naturally occurring views  satisfying the technical assumptions required to improve performance. Constructing randomized views is a cheap workaround that dramatically extends co-training's applicability.

\paragraph{Outline.} 
Section \S\ref{s:selectron} shows that discretizing standard models of synaptic dynamics and plasticity \cite{bb:12} leads to a small tweak on linear regression. Incorporating NMDA synapses into the model as a second view leads to a neuronal co-training algorithm. 

Section \S\ref{s:xnv} reviews recent work which translated the above observations about neurons into a learning algorithm. We introduce Correlated Nystr\"om Views (\cksny), a state-of-the-art semi-supervised learning algorithm that combines multi-view regression with random views via the Nystr\"om method \cite{mbb:13}.

Finally, \S\ref{s:rcc} returns to neurons and sketches preliminary work analyzing the benefits of randomized co-training in cortex.

\section{Modeling neurons as selective linear regressors}
\label{s:selectron}

The perceptron was introduced in the 50s as a simple model of how neurons learn \cite{Rosenblatt:1958fk}. It has been extremely influential, counting both deep learning architectures and support vector machines amongst its descendants. Unfortunately however, the perceptron and related models badly misrepresent important features of neurons. In particular, they treat outputs symmetrically ($\pm1$) rather than asymmetrically (0/1). This is crucial since spikes (1s) are more metabolically expensive than silences (0s). Further, by-and-large neurons only update their synapses after spiking. By contrast, perceptrons update their weights after every misclassification.

To build a tighter link between learning algorithms and neurocomputational models, we recently discretized standard models of neuronal dynamics and plasticity to obtain the selectron \cite{bb:12}. This section shows that, suitably regularized, the selectron is almost identical to linear regression. The difference is a \emph{selectivity} term -- arising from the spike/silence asymmetry -- that encourages neurons to specialize.

\paragraph{The selectron.}
Let $\cO = \{0,1\}^N$ denote the set of inputs and $\bR_{\geq0}^N$ the set of possible synaptic weights. Let $\langle\wt_j,\x\rangle_\vartheta:=\langle\wt_j,\x\rangle-\vartheta$, where threshold $\vartheta$ is fixed. Given $\x\in\cO$, neuron $n_j$ with synaptic weights $\wt_j$ outputs 0 or 1 according to 
\begin{equation}
	\label{e:threshold}
	f_{\wt_j}(\x) := H\big(\langle\wt_j,\x\rangle_\vartheta\big)
	\quad \text{where }H(\cdot)\text{ is the Heaviside function.}
\end{equation}
We model neuromodulatory signals via $p(\mu|\x)$ where $\mu\in \bR$, with positive values corresponding to desirable outcomes and conversely. Signals may arrive after a few hundred millisecond delay, which we do not model explicitly.

\begin{defn}[selectron]
	A threshold neuron is a selectron if its reward function takes the form
	\begin{equation}
		\label{e:reward}
		R(\wt_j) 
		= \sum_{\iota=1}^\ell \underbrace{\mu^\iota}_{\text{neuromodulators}}\cdot \underbrace{\langle\wt_j,\x^\iota\rangle_\vartheta}_{\text{excess current}}\cdot \underbrace{f_{\wt_j}(\x^\iota)}_{\text{selectivity}}
		=: \hat{\bE}_{(\x,\mu)}\left[\mu\cdot \langle\wt_j,\x\rangle_\vartheta\cdot f_{\wt_j}(\x)\right].
	\end{equation}	
\end{defn}

The reward function is continuously differentiable (in fact, linear) as a function of $\wt$ everywhere except at the kink $\langle\wt,\x\rangle=\vartheta$ where it is continuous but not differentiable. We can therefore maximize the reward via gradient ascent to obtain synaptic updates
\begin{equation}
	\label{e:stdp}
	\Delta \wt_{ij} \propto \mu_j(\x)\cdot \x_i \cdot f_{\wt_j}(\x).
\end{equation}
\begin{thm}[discretized neurons, \cite{bb:12}]\label{t:limit}
 	Discretizing Gerstner's Spike Response Model \cite{gerstner:02} yields \eqref{e:threshold}. Discretizing spike-timing dependent plasticity (STDP) \cite{song:00} yields \eqref{e:stdp}. 
Finally, STDP is gradient ascent on a reward function whose discretization is $\eqref{e:reward}$.
\end{thm}

If rewards are more common than punishments then \eqref{e:stdp} leads to overpotentiation (and eventually epileptic seizures). Neuroscientists therefore introduced a depotentiation bias into STDP. Alternatively, \cite{bb:12}, introduced an $\ell_1$-constraint on synaptic weights enforced during sleep. Below, we interpolate between the two approaches by replacing the constraint with a regularizer. 

\paragraph{Linear regression.}
We first recall linear regression to aid the comparison. Given data $(\x^\iota,y^\iota)_{\iota=1}^\ell$, regression finds parameters $\hat{\wt}$ minimizing the mean squared error
\begin{equation*}
	\hat{\wt}:= \argmin_\wt \hat{\bE}\Big[\frac{1}{2}\big(y -\langle\wt,\x\rangle\big)^2\Big].
\end{equation*}
One way to solve this is by gradient descent using 
\begin{equation}
	\label{e:nabla_MSE}
	\nabla_\wt 
	= \hat{\bE}\Big[ \big(y-\langle\wt,\x\rangle\big)\cdot\x\Big].
\end{equation}	

\paragraph{Selective linear regression.}
Now, if we regularize the reward in \eqref{e:reward} as follows
\begin{align}
	\label{e:slr_rew}
	\hat{\wt_j} := & \argmax_{\wt_j} \hat{\bE}\Big[
	\overbrace{\mu\cdot \langle\wt_j,\x\rangle_\vartheta\cdot f_{\wt_j}(\x)}^{\text{reward, \eqref{e:reward}}}
	-\frac{1}{2}\overbrace{\langle\wt_j,\x\rangle_\vartheta^2 \cdot f_{\wt_j}(\x)}^{\text{regularizer}}\Big]\\
	\notag
	= & \argmin_{\wt_j} \hat{\bE}\Big[ \frac{1}{2}\Big(
	\underbrace{\mu - \langle\wt_j,\x\rangle_\vartheta}_{\text{regression}}
	\cdot \underbrace{f_{\wt_j}(\x)}_{\text{selectivity}}
	\Big)^2\Big]
\end{align}
the result is \emph{selective} linear regression: a neuron's excess current predicts neuromodulation  \emph{when it spikes.} Since synaptic weights are non-negative, the neuron will not fire for $\x$ such that $\hat{\bE}[\mu|\x]<0$. Computing gradient ascent obtains
\begin{equation}
	\label{e:nabla_slr}
	\nabla_{\wt_j} 
	= \hat{\bE}\Big[
	\underbrace{\big(\mu-\langle\wt_j,\x\rangle_\vartheta\big)\cdot\x}_{\text{regression gradient, \eqref{e:nabla_MSE}}}
	\cdot \underbrace{f_{\wt_j}(\x)}_{\text{selectivity}}\Big]
	= \sum_{\{\iota|n_j \text{ spikes}\}}\Big( \mu^\iota-\langle\wt_j,\x^\iota\rangle_\vartheta\Big)\cdot\x^\iota.
\end{equation}
The synaptic updates are $\Delta\wt_{ij}\propto (\mu - \langle\wt_j,\x\rangle_\vartheta)\x_if_{\wt_j}(\x)$. In other words, if neuron $n_j$ receives spike $\x_i$ and produces spike $f_{\wt_j}(\x)$, then it modifies synapse $i\rightarrow j$ proportional to how much greater the rescaled neuromodulatory signal is than the excess current. 

The selectivity term in \eqref{e:nabla_slr} makes biological sense. There are billions of neurons in cortex, so it is necessary that they specialize. Neuromodulatory signals are thus ignored unless the neuron spikes -- providing a niche wherein the neuron operates.

\paragraph{Multi-view learning in cortex.}
There are two main types of excitatory synapse: AMPA and NMDA. So far we have modeled AMPA synapses which are typically feedforward and ``driving'' -- they cause neurons to initiate spikes. NMDA synapses differs from AMPA in that \cite{roelfsema:05}
\begin{itemize}
	\item they \emph{multiplicatively modulate} synaptic updates and
	\item they prolong, but do not initiate, spiking activity.
\end{itemize}

We model the two types of synaptic inputs as AMPA, $\x$, and NMDA, $\z$, views with synaptic weights $\vt$ and $\wt$ respectively. In accord with the observations above, we extend \eqref{e:slr_rew} by adding a \emph{multiplicative} modulation term. The NMDA view is encouraged to align with neuromodulators and is regularized the same as AMPA. The NMDA view has no selectivity term since it does not initiate spikes. 

Finally, we obtain a (discretized) neuronal co-optimization algorithm, which simultaneously attempts to maximize how well each view predicts neuromodulatory signals and aligns the two views on unlabeled data:
\begin{gather}
	\label{e:coreg1}
	\argmax_{\vt_j,\wt_j}\; 
	\hat{\bE}\Big[
	\overbrace{\mu\cdot\langle\vt_j,\x\rangle_\vartheta f_{\vt_j}(\x)
	+ \mu\cdot\langle\wt_j,\z\rangle}^{\text{reward for each view}}	
	\\
	\notag
	+ \underbrace{\alpha_{co}\langle\vt_j,\x\rangle_\vartheta\langle\wt_j,\z\rangle 
	}_{\text{multiplicative modulation}} 
	- \underbrace{\frac{\alpha_1}{2}\langle\vt_j,\x\rangle_\vartheta^2 f_{\vt_j}(\x)
	- \frac{\alpha_2}{2}\langle\wt_j,\z\rangle^2}_{\text{regularizer for each view}}
	\Big].
\end{gather}

The next section describes a semi-supervised regression algorithm inspired by \eqref{e:coreg1}.

\section{Correlated random features for fast semi-supervised learning}
\label{s:xnv}

This section translates the multiview optimization above into a workable learning algorithm.

\paragraph{From neurons to machine learning.}
We make three observations about cortical neurons:
\begin{itemize}
	\item Each neuron's inputs are preprocessed by millions of upstream neurons, providing a non-linear feature map analogous to the kernel trick \cite{scholkopf:02}.
	\item A neuron's synaptic contacts (not weights) are random to first approximation \cite{maass:02}.
	\item The  ``\emph{modulation} - \emph{regularizers}'' terms in \eqref{e:coreg1} resemble canonical correlation analysis \cite{hardoon:04}:
	\begin{equation}
		\label{e:corr_coeff}
		\text{CCA:}\quad
		\argmax_{\vt,\wt}
		\frac{\hat{\bE}\left[\langle\vt,\x\rangle \langle\wt,\z\rangle\right]}
		{\sqrt{\hat{\bE}[\langle\vt,\x\rangle^2]
		\cdot \hat{\bE}[\langle\wt,\z\rangle^2]}}
		= \frac{\text{``\emph{modulation}''}}{\text{``\emph{regularizers}''}}.
	\end{equation}
\end{itemize}

\begin{algorithm}[htp]	
\caption{\texttt{Correlated Nystr\"om Views (\cksny)}.\label{alg:xnv}}
	\algorithmicrequire\; Labeled data: $\{\x_\iota,y_\iota\}_{\iota=1}^\ell$ and unlabeled data: $\{\x_\iota\}_{\iota=\ell+1}^{\ell+u}$
  \begin{algorithmic}[1]
    \STATE {\bf \emph{Generate features.}} Sample $\hat{\x}_1,\ldots,\hat{\x}_{2M}$ uniformly from the dataset, compute the eigendecompositions of the sub-sampled kernel matrices $\hat{\Kt}^{(1)}$ and $\hat{\Kt}^{(2)}$ which are constructed from the samples $\{1,\ldots,M\}$ and $\{M+1,\ldots,2M\}$ respectively, and featurize the input: 
        \begin{equation*}
    		\quad \z^{(\mu)}(\x_\iota)\leftarrow \hat{\Dt}^{(\mu), -1/2} \hat{\Vt}^{(\mu)\top} \sq{\kappa(\x_\iota,\hat{\x}_1),\ldots, \kappa(\x_\iota,\hat{\x}_M) }\tr
			\text{ for }\mu\in\{1,2\}. 
        \end{equation*}

	\STATE {\bf \emph{Unlabeled data.}} Compute CCA bases $\Bt^{(1)}$, $\Bt^{(2)}$ and canonical correlations $\lambda_1,\ldots,\lambda_M$ for the two views and set
    $\bar{\z}_\iota \leftarrow \Bt^{(1)}\z^{(1)}(\x_\iota).$
    \STATE {\bf \emph{Labeled data.}} Solve
    \begin{equation*}
   \widehat{\wcca} := \argmin_{\wcca} \frac{1}{\ell} \sum_{\iota=1}^\ell \left(\wcca \tr \bar{\z}_\iota - y_\iota\right)^2 + \|\wcca\|^2_{CCA} + \gamma \|\wcca\|_2^2 ~.
    \end{equation*}    
  \end{algorithmic}
  \algorithmicensure\; $\widehat{\wcca}$
\end{algorithm}

The observations suggest neurons perform an analog of
\begin{itemize}
	\item kernelized multiview regression
	\item with random views and
	\item a CCA penalty.
\end{itemize}
To check these form a viable combination, we put the pieces together to develop Correlated Nystr\"om Views (\cksny), see  Algorithm~\ref{alg:xnv} and \cite{mbb:13}. Reassuringly, \cksny beats the state-of-the-art in semi-supervised learning \cite{ji:12}. 

\paragraph{Multiview regression.} The main ingredient in \cksny is multiview regression, which we now describe. Suppose the loss of the best regressor in each view is within $\epsilon$ of the best joint estimator.
\begin{equation}
	\label{e:multiview}
	\tag{A}
	\text{\underline{Multiview assumption:}}
	\quad\quad
	\lloss(f^{(\nu)}) - \lloss(f) \leq \epsilon
	\quad\text{for }\nu\in\{1,2\}.
\end{equation}
Introduce the canonical norm $\|\vt\|^2_{CCA}=\sum_i \frac{1-\lambda_i}{\lambda_i}(\bar{\vt}_i)^2$ where $\bar{\vt}_i$ are orthogonal solutions to \eqref{e:corr_coeff} with correlation coefficients $\lambda_i$.  Multiview regression is then
\begin{equation}
	\label{e:crr}
	\hat{\vt} := \argmin_{\vt} \hat{\bE}\Big[\big(y-\langle\vt,\x\rangle\big)^2 + \|\vt\|_{CCA}\Big].
\end{equation}
Penalizing with the canonical norm biases the estimator towards features that are correlated across both views (the signal) and away from features that are uncorrelated (the noise). Multiview regression is thus a specific instantiation of the general co-training principle that good regressors agree across views whereas bad regressors may not.

\begin{thm}[multiview regression, \cite{kakade:07}]
	The multiview estimator's error, \eqref{e:crr}, compared to the best linear predictor $f$, is bounded by
	\begin{equation*}
		\bE_{\text{data}}[\lloss(\langle\hat{\vt},\cdot\rangle)]-\lloss(f)
		\leq 5\epsilon +\frac{\sum_i \lambda_i^2}{n}.
	\end{equation*}
\end{thm}
According to the theorem, a slight increase in bias compared to ordinary regression is potentially more than compensated for by a large drop in variance. The reduction in variance depends on how quickly the correlation coefficients decay. For example, in the trivial case where the two views are identical, there is no benefit from multiview regression. To work well, the algorithm requires sufficiently different views (where most basis vectors are uncorrelated) that nevertheless both contain good regressors (that is, the few correlated directions are of high quality).

\paragraph{Randomization.}
To convert multiview regression into a general-purpose tool we need to construct pairs of views -- for any data -- satisfying two requirements. First, they should contain good regressors. Second, they should differ enough that their correlation coefficients decay rapidly.

A computationally cheap approach that does not depend on specific properties of the data is to generate random views. To do so, we used the Nystr\"om method \cite{williams:01} and random kitchen sinks \cite{rahimi:08}. A recent theorem of Bach implies Nystr\"om views contain good regressors in expectation \cite{bach:13} and a similar result holds for random kitchen sinks. Although there are currently no results on correlation coefficients across random views, recent lower bounds on the Frobenius norm of the Nystr\"om approximation suggest ``medium-sized'' views (a few hundred dimensional) differ sufficiently \cite{wang:13}. Empirical performance is discussed below.

\paragraph{Performance.}
Table~\ref{t:performance} summarizes experiments evaluating \cksny on 18 datasets, see \cite{mbb:13} for details. We do not report on random kitchen sinks, since they performed worse than Nystr\"om views. Performance is evaluated against kernel ridge regression (\texttt{KRR}) and a randomized version of \texttt{SSSL}, a semi-supervised algorithm with state-of-the-art performance \cite{ji:12}.\footnote{\texttt{SSSL}$_M$, the randomized version of \texttt{SSSL}, performs similarly to the original in a fraction of the runtime.}

\cksny typically outperforms \texttt{SSSL}$_M$ by between 10\% and 15\%, with about 30\% less variance. Both semi-supervised algorithms achieve less than half the error of kernel ridge regression.

\begin{table}
	\caption{Average performance of \cksny against \texttt{KRR} and randomized \texttt{SSSL} on 18 datasets.}
	\label{t:performance}
	\begin{tabular}{l r r r r r}
	\hline
	 & $\ell=100$ & $\ell=200$ & $\ell=300$ & $\ell=400$ & $\ell=500$ \\
	\hline
	Avg reduction in error vs \texttt{KRR} &  56\%  & 62\%  &  63\% &  63\% & 63\%  \\ 
	Avg $\downarrow$ in error vs \texttt{SSSL}$_{M/2M}$  &  11\%  & 16\%  &  15\% &  12\% & 9\%  \\ 
	Avg $\downarrow$ in standard-error vs \texttt{SSSL}$_{M/2M}$  &  15\%  & 30\%  &  31\% &  33\% & 30\% \\
	\hline
	\end{tabular}	
\end{table}

Importantly, \cksny is \emph{fast}. For $10,000$ points, \cksny runs in $<1s$ on a laptop whereas (unrandomized) \texttt{SSSL} takes $2300s$. For $70,000$ points, \cksny's runtime is $<30s$ whereas \texttt{SSSL} takes unfeasibly long.

\section{Randomized cortical co-training}
\label{s:rcc}

We describe work in progress on multiview neuronal regression.

\paragraph{Selective co-regularized least squares.}
The multiview optimization in \eqref{e:coreg1} can be rewritten, essentially, as co-regularized least squares \cite{sindhwani:05} with a selectivity term encouraging specialization:
\begin{align}
	\label{e:coopt}
	\argmin_{\vt_j,\wt_j}\; & \hat{\bE}_{\text{labeled}}\Big[
	\overbrace{\big(\alpha_1\cdot\langle\vt_j,\x\rangle_\vartheta f_{\vt_j}(\x) - \mu\big)^2}^{\text{selective regression on AMPA view}}
	+ \overbrace{\big(\alpha_2\cdot\langle\wt_j,\z\rangle - \mu\big)^2}^{\text{regression on NMDA view}}
	\Big]
	\\ \notag
	+\; & \hat{\bE}_{\text{unlabeled}}\Big[ \alpha_{co}\cdot 
	\underbrace{\big(\langle\vt_j,\x\rangle_\vartheta 
	-\langle\wt_j,\z\rangle\big)^2}_{\text{unsupervised co-regularization}}
	\Big].
\end{align}
The model closely resembles \cksny, with a penalty that is easier for neurons to implement.

The selectivity term in \eqref{e:coopt} ensures that neurons only predict the neuromodulatory signals when they spike. In other words, neurons have the flexibility to search for an AMPA view containing a good regressor. The NMDA weights are then simultaneously aligned with the neuromodulatory signal and the AMPA weights by the remaining two terms.

\paragraph{Guarantees.}
Theoretical guarantees for co-regularization are provided in \cite{rosenberg:07, sridharan:08}. As above, they depend on having good regressors in sufficiently different views. We briefly sketch how these apply.

The benefit from co-training depends on the extent to which co-regularizing with unlabeled shrinks the function space applied to the labeled data. In short, it depends on the Rademacher complexity of
\begin{equation*}
	{\mathcal J} := \Big\{(f+g)/2 :  \hat{\bE}\big[\alpha_1\cdot f(\x)^2 + \alpha_2\cdot g(\z)^2\big] + \hat{\bE}_{\textrm{unl}}\big[\alpha_{co}\cdot|f(\x)-g(\z)|^2\big]\leq 1\Big\}.
\end{equation*}
Denote the Gram matrices of the two views by
\begin{equation*}
	\mathbf{K}_1 = \left(\begin{matrix}
		A_{u\times u} & C_{u\times \ell} \\
		C^\intercal_{\ell\times u} & B_{\ell \times \ell}
	\end{matrix}\right)
	\quad
	\mathbf{K}_2 = \left(\begin{matrix}
		D_{u\times u} & F_{u\times \ell} \\
		F^\intercal_{\ell\times u} & E_{\ell \times \ell}
	\end{matrix}\right)
\end{equation*}
where $A_{ij} = \langle\x^i,\x^j\rangle$ and $D_{ij} = \langle\z^i,\z^j\rangle$ are dot-products of unlabeled data on the respective views, and other blocks are similarly constructed using mixed and labeled data. It is shown in \cite{rosenberg:07} that 
\begin{equation}
	\label{e:difference}
	\textrm{Radem}({\mathcal J})^2 \leq \frac{1}{\ell^2}\left(\frac{\textrm{tr}(B)}{\alpha_1} + \frac{\textrm{tr}(E)}{\alpha_2}
	-\sum_{i=1}^\ell \|C_{\bullet i} - F_{\bullet i}\|_{(\frac{\textrm{Id}}{\alpha_{co}} + \frac{A}{\alpha_1}+\frac{D}{\alpha_2})^{-1}}^2\right).
\end{equation}
The last term is of particular interest. Each column $C_{\bullet i}$ represents a labeled point in the first view by its dot-product with unlabeled points, and similarly for $F_{\bullet i}$ in the second view. The greater the \emph{difference} between the representations in the two views, as measured by \eqref{e:difference}, the lower the Rademacher complexity and the better the generalization bounds on co-training.

It remains to be seen how selective multiview regression performs empirically, and to what extent \eqref{e:difference} provides a good guide to the improvement in generalization performance of the original \emph{undiscretized} models.

\section{Discussion}

Co-training and randomization are two simple, powerful methods that complement each other well -- and which neurons appear to use in conjunction. No doubt there are more tricks waiting to be discovered. It is particularly intriguing that NELL, one of the more ambitious AI projects in recent years, uses various co-training strategies as basic building blocks \cite{carlson:10, carlson:10a, balcan:13}. Despite a large body of research on how humans learn categories and relations, it remains unknown how (or whether) individual neurons learn categories.

Although the results sketched here are suggestive, they fall far short of the full story. For example, since neurons learn online -- and only when they spike -- they face similar explore/exploit dilemmas to those investigated in the literature on bandits. It will be interesting to see if new (randomized) bandit algorithms can be extracted from models of synaptic plasticity.

\paragraph{Acknowledgements.}
I thank my co-authors Michel Besserve, Joachim Buhmann and Brian McWilliams for their help developing these ideas.


\end{document}